# IMAGE STITCHING WITH PERSPECTIVE-PRESERVING WARPING

Tianzhu Xiang, Gui-Song Xia, Liangpei Zhang

State Key Laboratory of Information Engineering in Surveying, Mapping,
and Remote Sensing, Wuhan University, Wuhan, China
Email: {tzxiang, guisong.xia, zlp62}@whu.edu.cn

**Commission III, WG III/3**

**KEY WORDS:** Image stitching, Image alignment, Perspective-preserving warping, Similarity transform, Projective transform

**ABSTRACT:**

Image stitching algorithms often adopt the global transformation, such as homography, and work well for planar scenes or parallax free camera motions. However, these conditions are easily violated in practice. With casual camera motions, variable taken views, large depth change, or complex structures, it is a challenging task for stitching these images. The global transformation model often provides dreadful stitching results, such as misalignments or projective distortions, especially perspective distortion. To this end, we suggest a perspective-preserving warping for image stitching, which spatially combines local projective transformations and similarity transformation. By weighted combination scheme, our approach gradually extrapolates the local projective transformations of the overlapping regions into the non-overlapping regions, and thus the final warping can smoothly change from projective to similarity. The proposed method can provide satisfactory alignment accuracy as well as reduce the projective distortions and maintain the multi-perspective view. Experiments on a variety of challenging images confirm the efficiency of the approach.

## 1. INTRODUCTION

Limited by the narrow field of view (FOV) of optical cameras, it is difficult to fully capture the entire scene with a single image. Image stitching is to combine multiple images with overlapping FOV into a single larger and seamless mosaic, which can extend the effective FOV of cameras. Image stitching has been widely studied in photogrammetry, remote sensing and computer vision (Brown and Lowe, 2007).

The key of image stitching is how to model the geometric transformations between multiple images with overlapping FOVs. Traditional image-stitching methods (Szeliski, 2005) try to estimate a global 2-dimensional transformation, typically homography, by corresponding points to align the input images. However, it is usually lack of flexibility to handle all types of scenes and motions. Generally speaking, the global transformation is accurate as long as some restrictive conditions are met, such as planar scenes or parallax free camera motion, for instance, the shooting location is fixed or only rotational motion is allowed. While in reality, especially in the days when smart phones and unmanned aerial vehicles have become more and more popular, the imaging conditions are beyond control, images are taken casually and often full of noticeable depth changes or complex scenes. Thus image-stitching algorithms with a single global transformation can no longer fit these data well, and often lead to unsatisfied visual artefacts, e.g. ghosting effects (misalignments), projective distortions (shape and perspective distortions).

There are often two kinds of methods (Zaragoza et al., 2013) to overcome the previously mentioned misalignment: (1) by estimating a more suitable image alignment model, or (2) by using advanced post-processing technologies after alignment such as compositing, blending. For second method, seam cutting (Chon et al., 2010) or multi-band blending (Brown and Lowe, 2007) is supposed to be a good choice, and some commercial stitching software, e.g. Microsoft Image Composite Editor (ICE) (Microsoft research, 2015) and Autostitch (Brown and Lowe, 2007), can provide advanced de-ghosting algorithms to remove the undesired artefacts. However, for stitching images taken under the uncontrolled imaging conditions, merely relying on the post-processing algorithms is inadequate, the obvious misalignments are hard to be eliminated by these methods. It has been reported that the precise model of alignment is vital important for image stitching (Lin et al., 2011). Then for the projective distortion problem, there are few methods at present. Most methods neglect the projective distortions, so some stitched area may be stretched or inconsistently enlarged, the perspective of the stitched image may be undesirably changed. This seriously affects the image stitching performance. This paper addresses the image alignment problem for image stitching, especially the projective distortions in image stitching.

Recently, some algorithms based on local warping models (Zhang and Liu, 2014) for accurate alignment have been proposed. Instead of estimating a single global homography, (Gao et al., 2011) proposed a dual-homography warping (DHW) model to stitch images. It first divides the scene into a ground plane and a distant plane, then estimates two transformations respectively, finally smoothly integrates the two by weight average. This method improves the visual effect of image stitching and works well for many simple scenes, but it is worth noticing that it is difficult to decide the number of the required planes for arbitrary scene.

Alternatively, (Lin et al., 2011) introduced the smoothly varying affine (SVA) method. This algorithm replaces a global affine transformation with a smoothly varying affine stitching field, so that it allows local geometric deformations. In contrast with other approaches, SVA is more flexible and tolerant to parallax, but it only has six degrees of freedom and thus it is not sufficient to achieve global projectivity.

In order to mitigate the drawbacks of SVA, (Zaragoza et al., 2013) presented the as-projective-as-possible warping (APAP). APAP uses a smoothly local projective transformation estimated by moving direct linear transformation (DLT), which aims to be globally projective while allowing local non-projective deviations. It has been reported to achieve outstanding performances on image alignment (Zaragoza et al., 2014). However, APAP may introduce shape and perspective distortions in non-overlapping regions, for the fact that it just smoothly extrapolates the projective transformation into these regions.

More recently, (Chang et al., 2014) proposed a shape-preserving half-projective (SPHP) warp for image stitching. From the overlapping regions to the non-overlapping regions, it applies three different continuous warps to achieve the smoothly change from projective to similarity. This algorithm can alleviate the shape distortion in the non-overlapping regions. However, it is hard to handle parallax. The combination of SPHP and APAP is an improvement version to handle parallax, but it is sensitive to parameter selection and it may introduce unsatisfactory local deformations.

Nevertheless, most methods only focus on the accuracy of alignment without considering distortions caused by projective transformation, especially perspective distortion. As argued in (Chang et al., 2014), using projective transformation only obtains a single-perspective stitched image and it will inevitably introduce perspective distortions. To solve this problem, the similarity transformation is introduced to tweak projective transformation to mitigate perspective distortions in this paper, because similarity transformation just consists of translation, rotation and uniform scaling and thus it doesn't bring about perspective distortion.

In this paper, we present a novel perspective-preserving warping, which integrates the multiple local homographies with global similarity transformation. More precisely, the proposed algorithm first divides the input image into grid mesh, and then estimates the local homographies by moving direct linear transformation for each mesh. By this way, an accurate alignment can be achieved in overlapping region. Subsequently, a global similarity transformation is introduced to compensate the perspective distortions in non-overlapping region by weighted integration with the local homographies. In addition, we also present a method to smoothly calculate the weight coefficients based on the analysis of the projective transformation. Experiments demonstrate that the proposed warping can not only be flexible to handle parallax, but also preserve the perspective well with less distortions.

The remainder of the paper is organized as follows. Section 2 details the proposed approach for image stitching. Section 3 then analyses and discusses the experimental results. Section 4 finally ends the paper with some concluded remarks.

## 2. THE PROPOSED ALGORITHM

This part details a complete presentation of the proposed algorithm. First we depict the estimation method of local warping by moving DLT. Then we describe the weighted combination of global similarity transformation and local warping.

### 2.1 Local warping

Given a pair of matching points $Y = [x\ y\ 1]^T$ and $Y' = [x'\ y'\ 1]^T$ in the target image $I$ and the reference image $I'$ respectively, a projective transformation $H$ for mapping $Y$ to $Y'$ can be obtained by:

$$\begin{bmatrix} x' \\ y' \\ 1 \end{bmatrix} = H \begin{bmatrix} x \\ y \\ 1 \end{bmatrix} = \begin{bmatrix} h_1 & h_2 & h_3 \\ h_4 & h_5 & h_6 \\ h_7 & h_8 & 1 \end{bmatrix} \begin{bmatrix} x \\ y \\ 1 \end{bmatrix} \quad (1)$$

The transformation matrix $H$ can be estimated by a group of corresponding points between $I$ and $I'$ using DLT. So the Eq. (1) can be rewritten by a cross product: $0_{3\times 1} = Y' \times HY$, that is

$$0_{3\times 1} = \begin{bmatrix} 0_{3\times 1} & -Y^T & y'Y^T \\ Y^T & 0_{3\times 1} & -x'Y^T \\ -y'Y^T & x'Y^T & 0_{3\times 1} \end{bmatrix} \begin{bmatrix} h_1 \\ \vdots \\ h_9 \end{bmatrix} \quad (2)$$

where the transformation $H$ is denoted in a $9\times 1$ vector. In the other $3\times 9$ matrix, in fact there are only two rows which are linearly independent. Given $N$ matching points $\{Y_i\}_{i=1}^N$, and $\{Y_i'\}_{i=1}^N$, $H$ can be estimated by

$$\hat{h} = \arg\min_h \sum_{i=1}^N \|a_i h\|^2 = \arg\min_h \|Ah\|^2, \ s.t. \|h\| = 1 \quad (3)$$

where $a_i$ is the two linearly independent rows, and $A$ is a $2N\times 9$ matirx. The solution is the the least significant right singular vector of $A$.

In APAP (Zaragoza et al., 2013), it first partitions the target image into grid meshes. For each mesh, it adopts a location dependent homography. So the local homography is estimated from the weighted problem

$$h_* = \arg\min_h \sum_{i=1}^N \|w_*^i a_i h\|^2, \ s.t. \|h\| = 1 \quad (4)$$

where $w_*^i$ denotes the influence of each pair of point conspondences on the $i^{th}$ gird. Assuming $x_*$ denotes the center of the $i^{th}$ gird, the weights are calculated as

$$w_*^i = \max\left(\exp\left(-\|x_* - x_i\|^2 / \sigma^2\right), \eta\right) \quad (5)$$

where $\sigma$ is the scale parameter, and $\eta \in [0\ 1]$ is used to avoid the numerical issues in estimation. From Eq. (5), the weight is high when the center point is closer to the matching points, and approximately equal when the center point is far from the matching points. So the estimated $H$ can fit the local structures.

As can be seen in Figure 4 (a) and (b), the overlapping regions are well aligned, but the stitched result suffers from the obvious shape and perspective distortions in the non-overlapping regions. In our algorithm, the similarity transformation is then employed to compensate the distortions, as detailed below.

## 2.2 Optimal similarity transformation

Similarity transform can be regarded as a combination of panning, zooming and in-plane rotation of a camera, which keeps the viewing direction unchanged, and thus it can preserve the perspective. If we can find a global similarity transformation that approximates the camera in-plane motion between the reference and target images, it can offset the influence of camera in-plane motion to some extent. While using all point correspondences for the estimation of similarity transformation like (Chang et al., 2014) is not reasonable because the overlapping regions may contains various image planes. From (Lin et al., 2015), the plane that is most parallel to the image projective plane at the focus length of the camera can be used to estimate the optimal similarity transformation by its point correspondences.

To find out point correspondences on this plane, RANSAC (Chin et al., 2012) is adopted to segment the matching points iteratively. Given the threshold $d$ (in our experiments, 0.01 is a good choice), RANSAC is used to find a group of points with largest inliers, and remove inliers, then repeat these steps on the left outliers until inliers are less than $\delta$, e.g. 50 inliers. For each group of inliers, a similarity transformation can be estimated, and the corresponding rotation angle can be calculated. The similarity transformation with smallest rotation is the one we preferred. The parameters mentioned above are selected by experience and experiments. Figure 1 shows an example of calculating the optimal global similarity transformation. Different colours denote different groups of point correspondences. Here the blue point group is selected to estimate the optimal similarity transformation whose rotation angle is the least.

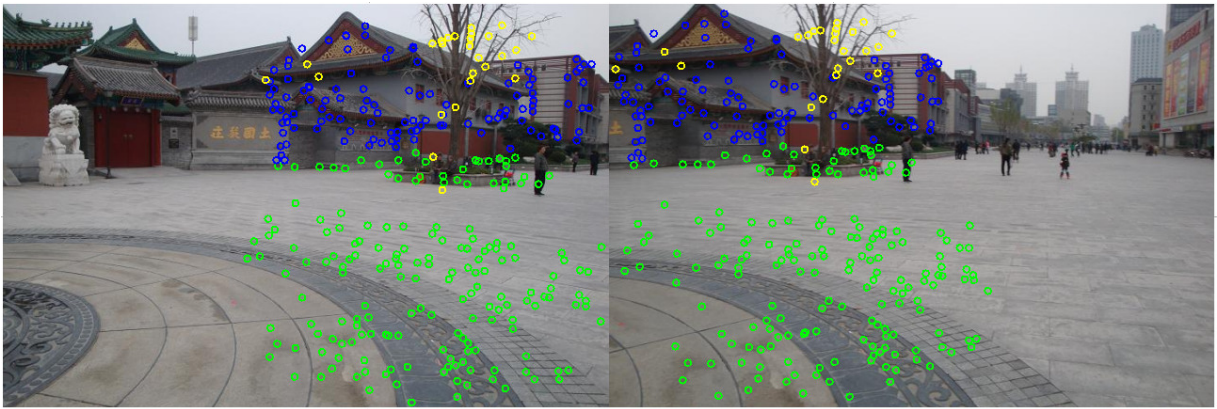

Figure 1. The optimal similarity transformation estimation

## 2.3 Weighted combination of similarity transformation

After obtaining the global similarity transformation, it can be used to adjust the local warping by weighted combination. For the smoothly transition, the whole image is taken into consideration to estimate the final warping. The weighted integration can be calculated as

$$H_i^{'} = \alpha H_i + \beta S \qquad (6)$$

where $H_i$ is the local warping in the $i^{th}$ grid. $H_i^{'}$ is the final local warping, and $S$ is the similarity transformation. $\alpha$ and $\beta$ are weight coefficients with the constraint $\alpha + \beta = 1$. Figure 2 shows the change of weights of local warping and similarity transformation for *Temple* images in Figure 3 (a). The pixel value denotes the weight coefficient. The greater the pixel value is, the higher the weight is. From Figure 2, the area is far from the overlapping regions, especially the distorted non-overlapping regions, it assigns a high weight for similarity transformation so that it can mitigate the distortions as much as possible, while for the area near the overlapping regions, it assigns a high weight for the local warping so as to ensure the accuracy of alignment.

Because the local warping in the overlapping regions is tweaked, the reference image should also take a corresponding warping to guarantee the alignment between the target image and the reference image. The warping for reference image is

$$T_i^{'} = H_i^{'} H_i^{-1} \qquad (7)$$

where $T_i^{'}$ is the local warping for the reference image in the $i^{th}$ grid.

Because distortions in the non-overlapping regions do not gradually change along the $x\text{-}axis$, the weight integration scheme presented in (Lin et al., 2015) is not an optimal solution. If the direction where the distortion just occurs along exists, the weight coefficients can be better estimated.

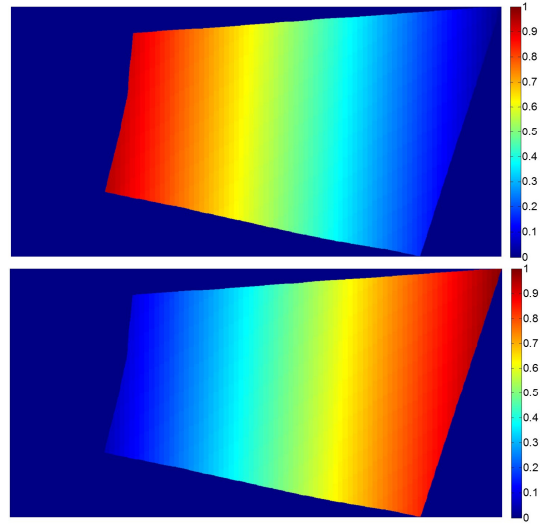

Figure 2. Weight map for local warping (top) and global similarity transformation (bottom)

## 2.4 Weight calculation

Early in (Chum et al., 2005), it applies the change of coordination to estimate the geometric error for the homography, and it can also help to reveal the distortion characteristics of projective transformation.

According to (Chum et al., 2005), the coordination $(x, y)$ of the target image is rotated to a new coordination system $(u, v)$. The relationship between them can be described as:

$$\begin{bmatrix} x \\ y \\ 1 \end{bmatrix} = \begin{bmatrix} \cos(\theta) & -\sin(\theta) & 0 \\ \sin(\theta) & \cos(\theta) & 0 \\ 0 & 0 & 1 \end{bmatrix} \begin{bmatrix} u \\ v \\ 1 \end{bmatrix} \quad (8)$$

So a new projective transformation $Q$ that transformations $(u, v)$ to $(x', y')$ is acquired which meets the formula below

$$\begin{bmatrix} x' \\ y' \\ 1 \end{bmatrix} = \begin{bmatrix} q_1 & q_2 & q_3 \\ q_4 & q_5 & q_6 \\ q_7 & q_8 & 1 \end{bmatrix} \begin{bmatrix} u \\ v \\ 1 \end{bmatrix} = \begin{bmatrix} h_1 & h_2 & h_3 \\ h_4 & h_5 & h_6 \\ h_7 & h_8 & 1 \end{bmatrix} \begin{bmatrix} \cos(\theta) & -\sin(\theta) & 0 \\ \sin(\theta) & \cos(\theta) & 0 \\ 0 & 0 & 1 \end{bmatrix} \begin{bmatrix} u \\ v \\ 1 \end{bmatrix} \quad (9)$$

Supposing $\theta = \arctan(h_8 / h_7)$, and combining with the above fomula, we can get $q_8 = -h_7 \sin(\theta) + h_8 \cos(\theta) = 0$. So $Q$ can be rewritten as

$$Q = \begin{bmatrix} q_1 & q_2 & q_3 \\ q_4 & q_5 & q_6 \\ -c & 0 & 1 \end{bmatrix} \quad (10)$$

where $c = \sqrt{h_7^2 + h_8^2}$. And then $Q$ can be decomposed as below:

$$\begin{bmatrix} q_1 & q_2 & q_3 \\ q_4 & q_5 & q_6 \\ -c & 0 & 1 \end{bmatrix} = \underbrace{\begin{bmatrix} q_1 + cq_3 & q_2 & q_3 \\ q_4 + cq_6 & q_5 & q_6 \\ 0 & 0 & 1 \end{bmatrix}}_{Q_a} \underbrace{\begin{bmatrix} 1 & 0 & 0 \\ 0 & 1 & 0 \\ -c & 0 & 1 \end{bmatrix}}_{Q_p} \quad (11)$$

where $Q_a$ is affine transformation and $Q_p$ is projective transformation. According to (Kimmel et al., 2011), the local scale change at point $(u, v)$ under the projective transformation is defined as the scale change of the first order matrix of $Q$ (i.e. Jacobin matrix of $Q$) at point $(u, v)$, so in terms of $Q_a$ and $Q_p$, the local scale change is calculated as below:

$$\det J(u,v) = \det J_a(u,v) \cdot \det J_p(u,v) = k_a \cdot \frac{1}{(1-cu)^3} \quad (12)$$

where det denotes the determinant, and $k_a$ is independent of $u$ and $v$. From the Eq. (12), the local scale change derived from $Q$ only relies on $u$, that is, the distortion of projective transformation $Q$ only occurs along the $u$-axis. (Chang et al., 2014) also expounds this character of projective transformation.

Obtaining $u$-axis, the image can be transformed to $(u,v)$ coordination system and the projective distortion can be compensated along $u$-axis. For simplicity, the centre of image is used as the origin of coordination, and the rotation transformation is applied on the reference image and the warped target image. Supposing $o$ is the origin point, the unit vector on $u$-axis denotes $\overrightarrow{ou} = (1, 0)$. For every grid mesh point $P_i$ (the $i^{th}$ centre of grid), $d_i$ is the projected length of vector $\overrightarrow{op_i}$ on vector $\overrightarrow{ou}$. And the maximum length among set $d$ corresponding projective point $P_{max}$ and the minimum length among set $d$ corresponding projective point $P_{min}$ can be obtained. So the weighting coefficients can be calculated as:

$$\beta = <\overline{P_{min}P_i} \cdot \overline{P_{min}P_{max}}> / |\overline{P_{min}P_{max}}| \qquad (13)$$

$$\alpha = 1 - \beta \qquad (14)$$

where $<\overline{P_{min}P_i} \cdot \overline{P_{max}P_{min}}>$ denotes the projective length of $\overline{P_{min}P_i}$ on $\overline{P_{max}P_{min}}$.

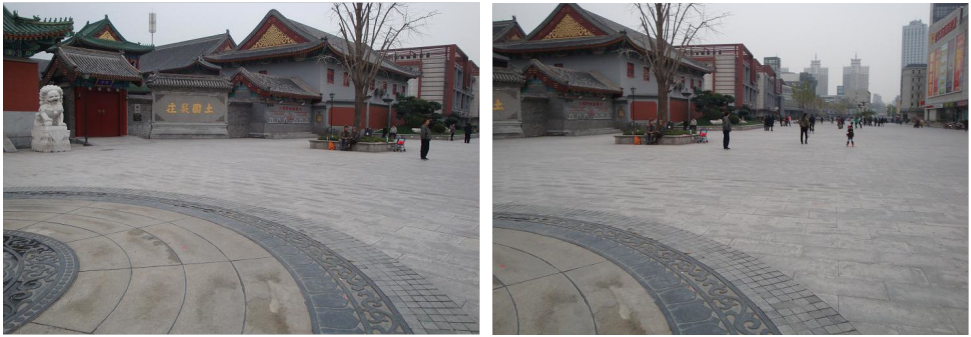

(a) Temple images

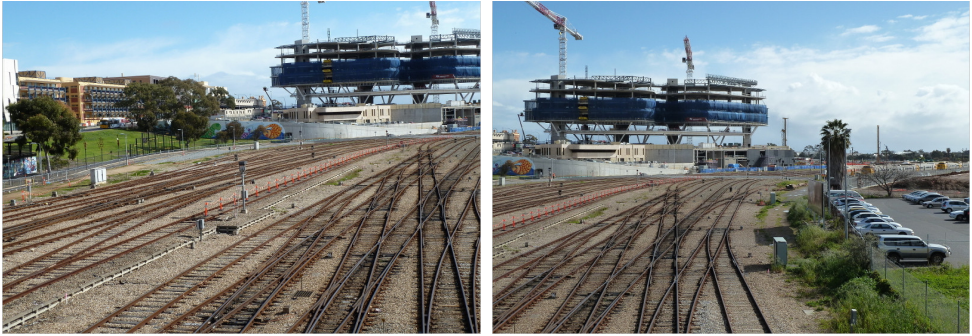

(b) Railtracks images

Figure 3. Original images for experiments

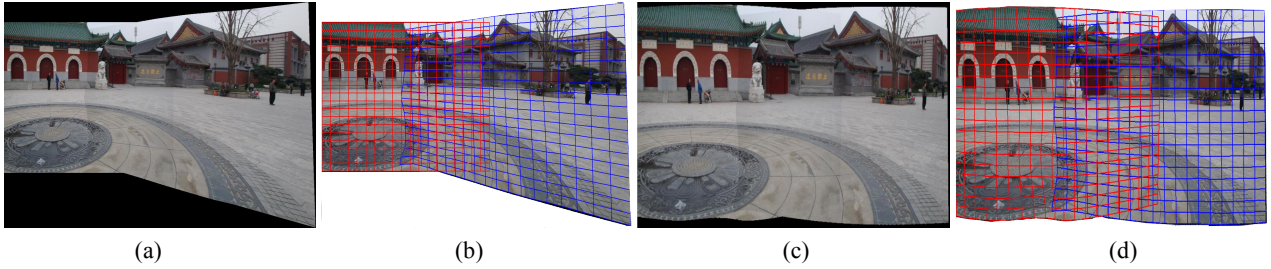

Figure 4. (a)-(b) APAP warping. (c)-(d) our warping.

## 3. EXPERIMENTS

In this part, a set of experiments are conducted on a range of challenging images captured casually. In order to evaluate the efficiency of our approach, we also compare it with the other algorithms, including ICE (Microsoft research, 2015), APAP (Zaragoza et al., 2013), SPHP with global homography (Chang et al., 2014), and SPHP with APAP (SPHP+APAP) (Chang et al., 2014).

To better compare the methods and reduce the interference, post-processing methods like blending or seam cutting detailed in (Szeliski, 2005) are avoided. The aligned images are simply blended by intensity average so that any misalignments remain obvious.

In our experiments, the key points in the reference image and target image are detected and matched by SIFT (Lowe, 2004). RANSAC (Chin et al., 2012) is used to remove mismatches. The parameters for other methods are set as the same suggested in each paper. The parameters of our method are a few, which are set as $\sigma = 8.5$, $\eta = 0.01$ in all our experiments.

Figure 3 illustrates the original images for experiments. Figure 3 (a) shows a pair of *Temple* images taken from different views. The scene contains large depth changes, and distinct multiple planes. Figure 3 (b) displays a pair of *Railtracks* images taken at different locations. The scene is complex and full of finely structures, such as tracks and steel structures. These taken conditions are great challenge for image stitching.

Figure 4 shows the comparison of APAP warping and the proposed warping. We can see that APAP in Figure 4 (a) and (b) uses a smoothly local projective warping that can allow local deviations to achieve a good alignment performance in the overlapping regions, while some severe shape distortion and perspective distortion occur in the non-overlapping regions, such as the ground bricks are stretched non-uniformly, the temple is undesirably enlarged. Whereas the results of our warping in Figure 4 (c) and (d) can guarantee the accurate alignment as well as gracefully alleviate distortions and maintain the multi-perspective view.

Figure 5 and 6 depict the stitching results on the *Temple* and *Railtracks* images. The results of each experiment are in the following order: ICE, APAP, SPHP, SPHP+APAP, and our approach. And we also highlight some special areas of each result. Red boxes show parallax errors or distortions, and green boxes show the satisfactory stitching result.

The ICE achieves a good visually performance to some extent. It maintains the perspective well and has less distortions in the non-overlapping regions. For the *Temple* images, the perspective and the shape of buildings are preserved. For *Railtracks* images, the palm tree is upright without tilt, the automobiles are without stretch. While to one's regret, ICE is not able to align the images satisfactorily, and thus it suffers from the obvious misalignments and ghosting effects, such as broken ground bricks in *Temple* result in Figure 5, and broken tracks in *Railtracks* result in Figure 6. This is because ICE applies the global homography for image stitching, its transformation model maybe inadequacy for the data.

Adopting local varying warping, APAP method can avoid ICE's drawbacks, and achieve impressive alignment performance in the overlapping regions with few artefacts. There are few misalignments for these images, whether the ground nearby or the temple in the distance in Figure 5, whether the complex rail tracks or the distant construction site in Figure 6. However, APAP cannot handle distortions well. It suffers from shape and perspective distortions in non-overlapping regions. In Figure 5, it is obviously that the building's shape in *Temple* result is no longer rectangular, and becomes the tilted quadrilateral. In Figure 6, the same defects can be seen, the palm tree is tilted and the automobiles are undesirably stretched and enlarged.

The SPHP and SPHP+APAP can mitigate the shape distortion and preserve the shape well, for instance, the buildings in *Temple* result remain rectangular, and the automobiles in *Railtracks* result keep the original appearance, because SPHP introduces the similarity transformation to offset these geometric distortions. Because SPHP adopts a global projective transformation, it also undergoes noticeable alignment errors, especially the roof of the temple in Figure 5, and the tower crane in Figure 6. With the help of APAP, SPHP+APAP method can greatly improve the alignment issue, and thus the roof of the temple and the tower crane are aligned accurately. However, SPHP+APAP method has two problems. First, some structures exist undesirable deformations, e.g. the temple behind the trees in Figure 5 and the building site in the distant in Figure 6. Then this method still exist perspective distortion and thus it cannot provide the fine perspective. In Figure 5, the second image is stitched with a global clockwise rotation, and the

*Railtracks* result in Figure 6 suffers from the same trouble. That is because the similarity transformation adopted in SPHP and SPHP+APAP is estimated by all point correspondences, it is not optimal to compensate distortions. This is particularly true if the scene is full of depth change and contains multiple distinct planes, just like Figure 3.

The last row of each experiment is the results of our method. The proposed method gracefully combines local warping and global similarity transformation. What's more, the estimated global similarity transformation is better for compensating the distortions than SPHP and SPHP+APAP. From the results of *Temple* in Figure 5 and *Railtracks* in Figure 6, the proposed method can successfully handle parallax, thus to attain the satisfactory alignment accuracy in the overlapping regions. Moreover, it can alleviate the projective distortions, especially the perspective distortion, so it can preserve the shape and perspective of each image well. From the comparison, our method achieves the best visual effects, and provides an impressive image stitching results.

## 4. CONCLUSION

This paper presents a perspective-preserving warping for image stitching. We observed that if the views of images do not differ purely by rotation or are not of a planar scene, these images are often difficult to align accurately in the overlapping regions, and are often troubled by distortions in the non-overlapping regions. The proposed approach gracefully combines the local warping with the global similarity transformation, which can handle parallax and distortions effectively. It weakly relies on the choice of parameters, and the suggested parameters can obtain a pleasant result for most situations.

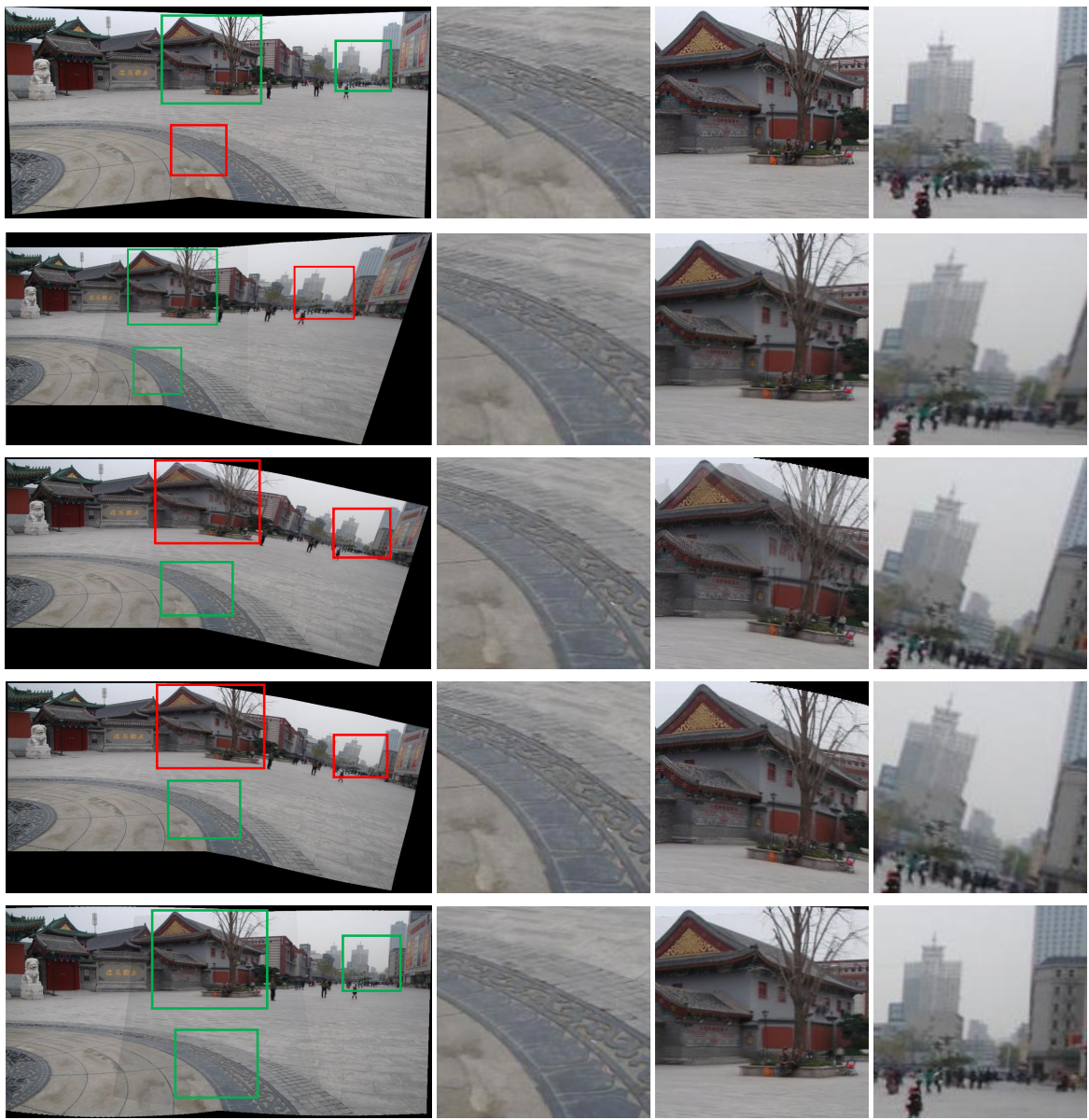

Figure 5. Comparison with the other image stitching algorithms on *Temple* images. From first row to last row, the results are: ICE, APAP, SPHP, SPHP+APAP, and our approach.

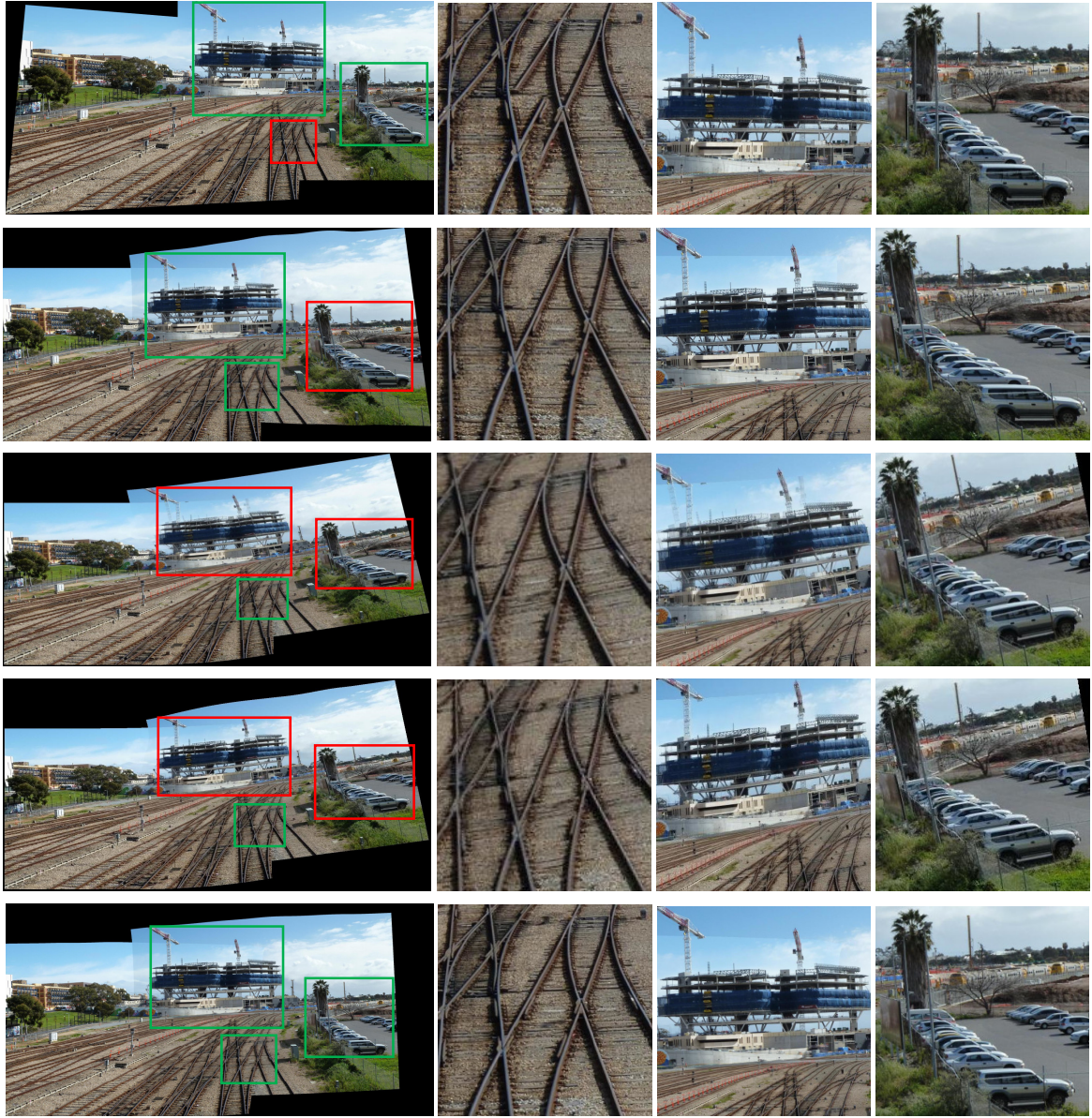

Figure 6. Comparison with the other image stitching algorithms on the *Railtracks* images. From first row to last row, the results are: ICE, APAP, SPHP, SPHP+APAP, and our approach.

The experimental analysis shows that the proposed method can achieve accurate alignment in the overlapping regions as well as reduce distortions and preserve perspective. It achieves the best stitching performance compared with the other methods. For the future work, it is of interest to investigate the use of more comprehensive local geometric transformation, such as that affine transformation induced by using local geometries (Xia et al., 2014).

**ACKNOWLEDGEMENT**

This research was supported by the National Natural Science Foundation of China under contract No. 91338113 and No. 41501462, and was partially funded by the Wuhan Municipal Science and Technology Bureau, with Chen-Guang Grant 2015070404010182.